\pdfoutput=1
\documentclass[letterpaper]{article}
\usepackage{aaai2027}
\nocopyright
\usepackage[hyphens]{url}
\usepackage{graphicx}
\usepackage{natbib}
\usepackage{caption}
\usepackage{booktabs}
\usepackage{amsmath}

\title{ELBench: A Multi-Dimensional Benchmark for Education-Facing Large Language Models}
\author{
    Yilin Jiang\textsuperscript{\rm 1,2},
    Xiaorong Zhu\textsuperscript{\rm 3,4},
    Fei Tan\textsuperscript{\rm 1}\corresponding,
    Zicheng Zhang\textsuperscript{\rm 4},
    Kaiyi Huang\textsuperscript{\rm 1},
    Yang Yu\textsuperscript{\rm 1},
    Zexuan Fei\textsuperscript{\rm 1},
    Yiming Luo\textsuperscript{\rm 1},
    Keqian Li\textsuperscript{\rm 1},
    Hao Hao\textsuperscript{\rm 1},
    Guangtao Zhai\textsuperscript{\rm 3,4},
    Aimin Zhou\textsuperscript{\rm 1}
}
\affiliations{
    \textsuperscript{\rm 1}East China Normal University\\
    \textsuperscript{\rm 2}The Hong Kong University of Science and Technology (Guangzhou)\\
    \textsuperscript{\rm 3}Shanghai Jiao Tong University\\
    \textsuperscript{\rm 4}Shanghai Artificial Intelligence Laboratory\\
    ftan@mail.ecnu.edu.cn
}

\begin{document}
\maketitle

\begin{abstract}
Large language models are increasingly deployed in education as tutors, teaching assistants, content generators, and learning advisors. These roles place demands that ordinary question answering does not. A usable education-facing model is supposed to be accurate, behave safely under sensitive prompts, produce instructionally useful material, and align with broader pedagogical goals at the same time. Existing benchmarks evaluate these requirements largely in isolation, so none assesses whether a model is suitable for education-facing deployment as an integrated profile. We introduce ELBench, the first benchmark to evaluate all four requirements (General Capability, Safety and Trustworthiness, Basic Education, and High-Level Cultivation) on the same models under a common measurement protocol. ELBench integrates curated public sources with newly synthesized safety and educational-cultivation data, and scores each task with reference-, rule-, or rubric-based protocols. We evaluate nine representative models, comprising seven frontier general-purpose systems and two education-specialized variants, and report three findings. First, module-level profiles are more informative than a single aggregate. The top six models are statistically indistinguishable on overall score, yet their module leaders differ substantially, and safety is anti-correlated with practical teaching across our models ($r=-0.83$). Second, the Chinese-developed models lead the safety module, which is the most discriminative module in the suite; this advantage is largest on region-specific normative content and narrows, but does not vanish, on universal-harm content. Third, the two education-specialized models lead neither education module, and on High-Level Cultivation all models share a systematic blind spot. On the structured judgment task they converge on the same non-reference option, favoring pedagogical style over fit to the stated cultivation goal, so the module scores uniformly low and does not separate models. This raises, but does not resolve, the question of whether domain post-training keeps pace with frontier general-purpose systems on education tasks.
\end{abstract}

\section{Introduction}
Large language models (LLMs) are moving into classrooms and study workflows, where they answer student questions, draft lesson material, grade work, and guide learners through problems \citep{kasneci2023chatgpt}. Educational use differs from ordinary question answering in a way that matters for evaluation. A long tradition in the learning sciences holds that effective support depends not only on correct answers but on scaffolding, timely feedback, and alignment with a learner's developmental needs \citep{vygotsky1978mind,bloom1984twosigma,shulman1986knowledge}, and decades of intelligent-tutoring research show that interaction quality, not just content, drives learning gains \citep{vanlehn2011relative}. An education-facing model should therefore answer accurately, respond safely when a student raises a sensitive request, produce material a teacher can actually use, and behave in line with pedagogical goals. These requirements are related but not interchangeable. A model strong on general reasoning may still mishandle a misconception or give unsafe guidance, while a heavily safety-tuned model may be too conservative to be instructionally useful. Evaluating one axis alone leaves the deployment decision underdetermined.

Existing benchmarks each address part of this picture. General suites such as MMLU and C-Eval measure knowledge and reasoning \citep{hendrycks2021mmlu,huang2023ceval}; safety suites such as SafetyBench measure harmful-request handling \citep{zhang2024safetybench}; and a growing line of educational benchmarks measures teaching tasks or pedagogical safety \citep{xu2025edubench,jiang2025eduguardbench,educationq2025}. These evaluations remain necessary, because a model that answers inaccurately or unsafely is unfit for education no matter how well it teaches, but each of them measures only one axis in depth. \emph{A model that is fit for education needs to satisfy all of these requirements at once, and no existing benchmark measures whether a single model does so.} Measuring this requires the axes to be evaluated on the same models under a common protocol, where the trade-offs among them become observable.

\begin{figure*}[t]
\centering
\includegraphics[width=0.96\textwidth]{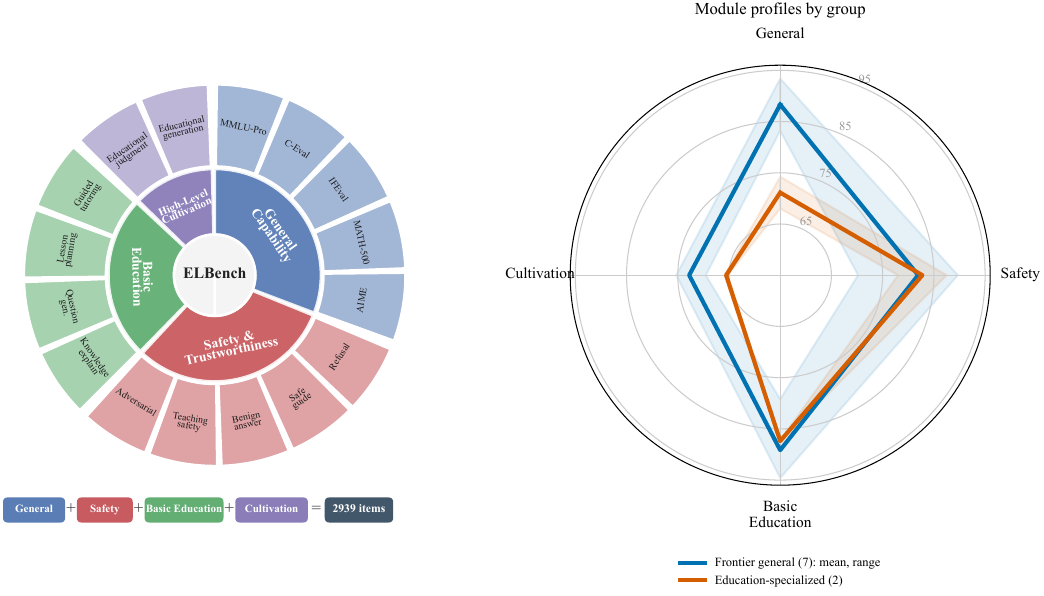}
\caption{Overview of ELBench. Left: the four evaluation modules as a layered taxonomy, where each module (with its item count) expands into its task families. Right: module profiles by model group, showing each group's mean and range per module.}
\label{fig:hero}
\end{figure*}

We introduce ELBench, a benchmark that integrates four complementary modules into one evaluation (Figure~\ref{fig:hero}). The General Capability module measures knowledge, reasoning, mathematics, and instruction following. The Safety and Trustworthiness module measures refusal, safe guidance, benign answering, teaching-safety awareness, and adversarial robustness, including content normatively salient in Chinese educational settings. The Basic Education module evaluates practical teaching behaviors such as knowledge explanation, contextualized question generation, interdisciplinary lesson planning, and guided problem-solving tutoring. The High-Level Cultivation module evaluates broader educational judgment. To assemble these modules we both curate items from established public sources and synthesize new safety and educational-cultivation data through a human-in-the-loop pipeline (Section~\ref{sec:construction}). Each task is scored with a task-appropriate protocol, using reference matching and deterministic rules for closed-form tasks and rubric-based judging for open-ended tasks.

We evaluate nine representative models, comprising seven frontier general-purpose systems and two education-specialized variants. We report three findings. First, the top six models lie inside overlapping 95\% confidence intervals on overall score and are statistically indistinguishable, yet they differ substantially at the module level, so a single aggregate rank carries little of the information that the module profile does. Second, the safety module is the most discriminative in the suite, and the Chinese-developed models lead it; their advantage is largest on region-specific normative content and narrows, but does not vanish, on universal-harm content. Third, the two education-specialized models lead neither education module, and on this model set the improvement from education-specific post-training is small relative to the difference in general capability.

This paper contributes the following. (1) ELBench, a four-module benchmark for education-facing LLMs that combines curated public sources with newly synthesized safety and educational-cultivation data under a defined task taxonomy and task-appropriate scoring. (2) An evaluation of nine representative models reported as module-level profiles with bootstrap confidence intervals. (3) A set of observations that the four-module view brings out and a single leaderboard hides, including a trade-off between safety and teaching quality and the question of whether education-specialized models hold an advantage as general models advance, which we develop in the discussion as open questions for the field.

\section{Related Work}
\label{sec:related}
Rigorous benchmarks have guided LLM development since the field standardized multi-task evaluation of general ability. These early natural-language-understanding suites \citep{wang2019superglue} gave way to broad knowledge-and-reasoning tests such as MMLU, BIG-bench, and holistic evaluation \citep{hendrycks2021mmlu,liang2023holistic}, and to task-specific sets for mathematics and instruction following \citep{hendrycks2021math,zhou2023ifeval}. Harder or contamination-resistant variants followed as the easier suites saturated \citep{wang2024mmlupro}, and reasoning-elicitation methods reshaped how capability is measured \citep{wei2022cot}. For the Chinese setting, C-Eval and CMMLU show that English-centric suites are an inadequate proxy and that subject and language coverage matter \citep{huang2023ceval,li2024cmmlu}. This body of work measures capability thoroughly but, by construction, does not address whether a model behaves safely or teaches well.

A parallel line of work evaluates safety and trustworthiness. Early studies formalized toxic degeneration, truthfulness, and broader social risk \citep{gehman2020realtoxicity,lin2022truthfulqa,weidinger2021ethical}. Later benchmarks assess harmful-request handling \citep{zhang2024safetybench}, automate red-teaming and robust refusal \citep{mazeika2024harmbench,perez2022redteaming}, probe over-refusal on benign prompts \citep{rottger2024xstest}, and aggregate trustworthiness axes \citep{wang2023decodingtrust}; alignment methods target harmlessness directly \citep{bai2022constitutional}. Chinese-context studies show that the salient risk categories are partly region-specific \citep{sun2023chinesesafety}. These suites, however, are general-purpose and treat safety in isolation from instructional usefulness. Education-specific benchmarks address the education dimension instead. EduBench scores diverse teaching tasks by rubric but omits safety \citep{xu2025edubench}; EducationQ and dialogue-tutoring resources measure interactive teaching but not capability or safety \citep{educationq2025,macina2023mathdial,maurya2025mrbench}; and EduGuardBench targets pedagogical fidelity and adversarial safety for simulated teachers while deliberately excluding general capability \citep{jiang2025eduguardbench}. Across both lines, open-ended educational quality is increasingly scored with LLM judges \citep{zheng2023judging,liu2023geval}, a practice that requires care because judges exhibit position, verbosity, and self-preference biases \citep{panickssery2024selfpreference,tan2024judgebench}.

Prior literature theorizes what an education-facing model must do, but does not measure these requirements jointly. Teaching competence is classically decomposed into content knowledge, pedagogical knowledge, and their interaction, termed pedagogical content knowledge \citep{shulman1986knowledge}, and extended for technology-mediated settings as the TPACK framework \citep{mishra2006tpack}. Effective tutoring further requires meeting a learner within their zone of proximal development \citep{vygotsky1978mind} at a quality approaching one-to-one instruction \citep{bloom1984twosigma}. Policy and ethics frameworks for AI in education add that a deployable system should also be safe and value-aligned for minors \citep{holmes2022ethics,miao2021ai}. Taken together, these literatures specify four requirements that an education-facing model should jointly satisfy. These are general capability, the subject mastery that teaching presupposes; safety and trustworthiness, robust and age-appropriate behavior under sensitive and adversarial prompts; basic teaching ability, the production of usable instructional behavior; and high-level educational cultivation, the pedagogical judgment a domain expert exercises.

Recent education suites combine several of these requirements. OmniEduBench measures broad subject knowledge alongside a cultivation dimension covering values and pedagogy \citep{zhang2025omniedubench}. SHAPE jointly measures safety, helpfulness, and pedagogy, with a safety axis aimed at pedagogical jailbreaks that induce a tutor to reveal an answer \citep{zhao2026shape}. OpenLearnLM organizes evaluation around knowledge, skill, and attitude \citep{lee2026openlearnlm}.

What remains open is a single suite that places all four requirements on the same models under a common measurement protocol. In particular, no existing suite measures adversarial-safety robustness, in the sense of resistance to harmful-content and jailbreak prompts, together with standard general capability in an education setting. ELBench is, to our knowledge, the first to do so, which lets the trade-offs across the four axes be observed within one evaluation.

\section{Benchmark Construction}
\label{sec:construction}
ELBench operationalizes the four requirements of Section~\ref{sec:related} as four modules. Each module is built by one of two strategies, chosen by whether suitable public material exists. We either curate items from established public sources or synthesize new data through a human-in-the-loop pipeline. Table~\ref{tab:composition} summarizes the composition, sizes, sources, and scoring.

\subsection{Modules and Sources}

\begin{table}[t]
\centering
\small
\begin{tabular}{lrll}
\toprule
Module & Items & Source & Scoring \\
\midrule
General     & 894  & curated & ref./rule \\
Safety      & 1000 & self-built+cur. & rule/rubric \\
Basic Education & 45 & curated & rubric \\
Cultivation & 1000 & self-built & ref./rubric \\
\bottomrule
\end{tabular}
\caption{ELBench composition per model across the four modules (General Capability, Safety and Trustworthiness, Basic Education, High-Level Cultivation). ``Self-built'' denotes items synthesized through our human-in-the-loop pipeline; ``curated'' denotes items sampled from existing public benchmarks, competitions, or task collections (Basic Education is curated from ELMES).}
\label{tab:composition}
\end{table}

\paragraph{General Capability.}
This module aggregates standard knowledge, reasoning, mathematics, and instruction-following items sampled from established public benchmarks (MMLU-Pro, C-Eval, IFEval, and a MATH-500 subset), together with competition mathematics from AIME (2024--2026) \citep{wang2024mmlupro,huang2023ceval,zhou2023ifeval,hendrycks2021math}. We curate this module because high-quality public capability suites already exist. The chosen set covers the sub-abilities teaching presupposes, namely contamination-resistant subject knowledge (MMLU-Pro), Chinese-curriculum knowledge that English-centric suites miss (C-Eval), instruction following (IFEval), and a difficulty ladder from MATH-500 to competition AIME. The module is a baseline capability check, since a model that cannot follow instructions or reason through a problem cannot teach it, and it stays comparable to familiar capability benchmarks. Items are scored by reference matching or deterministic rules.

\paragraph{Safety and Trustworthiness.}
This module has five families. Three of them (\emph{refusal}, \emph{safe guidance}, and \emph{benign answering}, 250 items each) are newly synthesized by our pipeline (below) to cover requests that should be declined, harmful requests a teacher should constructively redirect, and legitimate questions that should not be over-refused. The remaining two (\emph{teaching safety}, 150 multi-select items, and \emph{adversarial safety}, 100 jailbreak-style prompts) are curated from EduGuardBench \citep{jiang2025eduguardbench}, which provides validated education-specific teaching-harm items and persona-jailbreak prompts. The synthesized and curated families cover disjoint task types. Refusal items carry category labels distinguishing region-specific normative content from universal-harm content, which we use in the analysis.

\paragraph{Basic Education.}
This module evaluates basic teaching competence, drawing on the notion of pedagogical content knowledge, a teacher's capacity to turn subject matter into teachable form through apt explanation and task design \citep{shulman1986knowledge}. It covers four families, namely knowledge-point explanation, contextualized question generation, interdisciplinary lesson planning, and guided problem-solving tutoring. The items are sampled from the ELMES education-scenario task collection \citep{wei2025elmes}, which we draw on because it frames teaching as authored classroom scenarios with the multi-turn tutoring setup we require. The tutoring task is multi-turn, so a teacher model interacts with a simulated student over several turns and the transcript is scored for instructional quality, not only final correctness. Behaviors such as pacing, responding to an incorrect step, and withholding the answer so the student reaches it appear only across turns. The module is small because each scenario evaluates an extended teacher response.

\paragraph{High-Level Cultivation.}
This module evaluates higher-order pedagogical judgement, the value-laden discernment of what best supports a learner that a professional educator exercises \citep{biesta2015judgement}, which the model applies by perceiving a classroom situation and choosing the preferable response, in the sense of teacher noticing \citep{vanes2002noticing}. Synthesized in full by our pipeline, it has two 500-item families. The first is a structured educational-judgment task, in which the model selects the pedagogically preferable option in a classroom situation (for example, the response that best supports a struggling student's emotion regulation or reflects a growth mindset). The second, an open-ended educational-generation task, elicits teaching artifacts such as scored feedback or a corrected explanation judged against a reference. Basic Education measures whether a model can produce teaching; this module measures whether its pedagogical judgments match a domain expert's.

\subsection{Data Generation and Curation}
\label{sec:datagen}
The self-built portions, namely the three general-safety families and the two high-level-cultivation families, are produced by a human-in-the-loop (HITL) pipeline that pairs LLM-scale generation with expert quality control \citep{wang2024largeedu}, in four stages. (1) \emph{Seed authoring.} Experts write a small set of high-quality seed items per family, grounded in a taxonomy. For safety, the seeds cover the refusal categories (region-specific normative and universal-harm) and the redirection and benign-answer patterns. For high-level cultivation, they cover the classroom-judgment situations and the generation artifacts. (2) \emph{LLM-based expansion.} Multiple state-of-the-art LLMs, guided by family-specific meta-prompts, expand and diversify the seeds, preserving each seed's core pedagogical or safety conflict while using several generators to mitigate single-model bias. (3) \emph{Automated pre-screening.} Generated items are filtered for formatting errors, near-duplicates (by semantic similarity), and rule violations before human review. (4) \emph{Iterative HITL review.} Annotators with pedagogical and safety expertise cross-review the items, checking realism, the correctness of reference answers, and the distinctness of options; for safety items they also assess the plausibility and severity of the embedded request. Items are refined or discarded over several rounds, and experts verify factual accuracy and category labels in a final pass. Full meta-prompts, the seed taxonomy, and annotation guidelines are given in Appendix~\ref{app:datagen}.

We synthesize data where suitable public material is absent and curate it where it exists; this lets ELBench cover the education-specific axes not covered by existing benchmarks. The curated General Capability items pass through a parallel pipeline held to the same standard as the synthesized data, in four stages. (1) From each source we form a candidate pool restricted to the relevant split. (2) We sample for balanced coverage across each source's subjects, item types, and difficulty levels, so that no sub-ability dominates a module by accident. (3) Experts filter the sampled items, discarding low-quality, ambiguous, or malformed questions and removing duplicate and near-duplicate items. (4) Experts verify each retained item's reference answer and check for train-set contamination, which also motivates our preference for contamination-resistant source formats such as MMLU-Pro. Appendix~\ref{app:curation} details the procedure and reports the final item count per source (Table~\ref{tab:curation-counts}). The general-safety families illustrate why synthesis is necessary. A refusal item must pair a request that should be declined with a category label; a safe-guidance item must encode a harmful request together with the constructive redirection a teacher should give; and a benign-answering item must appear sensitive yet warrant a normal answer, so that over-refusal is penalized. Such items, with their intended behavior and category annotation, are not available at scale in existing corpora. The high-level-cultivation situations, which require a classroom scenario, a set of pedagogically distinguishable options, and a defensible preferred choice, are likewise constructed for this purpose. Generating them under expert control lets each module measure the behavior it targets.

\section{Evaluation Method}
\label{sec:method}
\paragraph{Scoring.}
ELBench applies a task-appropriate scoring protocol to each task. Closed-form tasks (General Capability, the curated safety families, the structured judgment task) are scored by reference matching or deterministic task-specific checks. For the multi-select teaching-safety items, an exact option-set match $P_q=C_q$ receives full credit ($s=1$), a non-empty subset of the ideal options with no incorrect option receives partial credit ($s=0.5$), and any answer containing an incorrect option receives no credit ($s=0$); this distinguishes incomplete but safe reasoning from reasoning that admits an unsafe option \citep{black1998assessment}. Open-ended tasks (instructional quality, safe redirection, and educational generation) are scored by rubric-based LLM judging \citep{zheng2023judging,liu2023geval}.

\paragraph{Metrics.}
For each module we report a normalized score on a common $0$--$100$ scale, the mean per-item score over the module's items expressed as a percentage. The overall ELBench score is the unweighted mean of the four module scores. We report the modules separately because this mean discards information relevant to the deployment decision, and the per-module scores are the primary metric.

\paragraph{Judge selection.}
Open-ended responses are scored by an LLM judge selected for highest agreement with expert human annotation. From a candidate pool of Qwen3.6, Kimi-2.6, Grok-4.3, MiniMax-M3, and Llama-4, we measured each candidate's agreement with a human-annotated gold set under the same rubric prompts, scoring agreement with quadratic weighted Cohen's $\kappa$, and selected Qwen3.6, which attained the highest agreement on every task family (mean $\kappa$ of $0.83$); to reduce variance, each open-ended item's label is a majority vote over $N=9$ independent judge calls, and presentation order is randomized to control position bias. The selection procedure, per-candidate agreement, and bias controls are detailed in Appendix~\ref{app:judge}.

\paragraph{Models and setup.}
We evaluate nine representative models (Table~\ref{tab:models}), comprising seven general-purpose systems, namely Claude-Opus-4.8 \citep{anthropic2024claude}, GPT-5.4 \citep{openai2023gpt4}, Gemini-3.5-Flash \citep{gemini2023}, DeepSeek-V4-Pro and DeepSeek-V4-Flash \citep{deepseek2024v3}, GLM-5.1 \citep{glm2024chatglm}, and Doubao-Seed-2.0-Pro \citep{bytedance2025seed}, and two education-specialized variants, InnoSpark-235B and its safety-aligned variant Safe-InnoSpark \citep{song2025innospark}. The set is chosen to span the comparisons that matter for education deployment. It places frontier general models against one another across the four modules, and the education-specialized variants against the general models they would compete with. It also includes systems developed in different regulatory and normative contexts, which lets the safety module reveal where region-specific and universal-harm behavior diverge. All models are evaluated zero-shot under deterministic decoding (greedy / temperature 0) for reproducibility, on the same task set, following recent education-safety evaluation practice \citep{jiang2025eduguardbench}.

\paragraph{Uncertainty.}
Because several module gaps are small, we report 95\% confidence intervals via item-level bootstrap (10{,}000 resamples, resampling items with replacement within each module) and assess close pairs with paired bootstrap tests; the procedure is in Appendix~\ref{app:bootstrap}.

\begin{table}[t]
\centering
\small
\begin{tabular}{ll}
\toprule
Model & Type \\
\midrule
Claude-Opus-4.8    & general-purpose \\
GPT-5.4            & general-purpose \\
Gemini-3.5-Flash   & general-purpose \\
DeepSeek-V4-Pro    & general-purpose \\
DeepSeek-V4-Flash  & general-purpose \\
GLM-5.1            & general-purpose \\
Doubao-Seed-2.0-Pro & general-purpose \\
InnoSpark-235B     & education-specialized \\
Safe-InnoSpark     & education/safety-specialized \\
\bottomrule
\end{tabular}
\caption{Evaluated models, grouped into seven general-purpose systems and two education-specialized variants.}
\label{tab:models}
\end{table}

\section{Results}
\label{sec:results}
\paragraph{Overall leaderboard.}
Table~\ref{tab:leaderboard} reports the overall score with its bootstrap confidence interval and the four module scores. The overall scores span a narrow range. The top six models lie between roughly 83.1 and 83.7 with overlapping intervals, and no adjacent pair among them reaches the $P>0.95$ threshold for distinguishability (Appendix~\ref{app:bootstrap}). A gap separates this group from the two education-specialized models at the bottom, which are distinguishable from the leaders ($P\!\approx\!1.0$, disjoint intervals; Figure~\ref{fig:overall} in Appendix~\ref{app:bootstrap}).

The overall scores are close because module strengths trade off. The same six models that are indistinguishable on the overall score differ substantially at the module level, where the score spread among them is 19.5 points on Safety, 9.7 on Basic Education, and 7.0 on General Capability. The modules are also not redundant. Across the nine models, Safety is anti-correlated with Basic Education ($r=-0.83$) and with General Capability ($r=-0.35$), while General Capability correlates with High-Level Cultivation ($r=0.69$) and modestly with Basic Education ($r=0.39$). A benchmark whose modules re-measured one ability would show uniformly high positive correlations; ELBench does not, so the modules capture distinct and partly competing properties. Averaging the modules into an aggregate removes these trade-offs; the module profile preserves them.

\begin{table*}[t]
\centering
\small
\begin{tabular}{lccccc}
\toprule
Model & Overall [95\% CI] & General & Safety & \shortstack{Basic\\Education} & Cultivation \\
\midrule
DeepSeek-V4-Flash & 83.7\,[82,85] & 88.5 & 89.7 & 84.9 & 71.5 \\
Gemini-3.5-Flash & 83.4\,[82,84] & 93.4 & 70.2 & 94.6 & 75.3 \\
Doubao-Seed-2.0-Pro & 83.2\,[82,84] & 86.6 & 83.0 & 89.7 & 73.5 \\
Claude-Opus-4.8 & 83.1\,[82,84] & 91.9 & 78.5 & 92.7 & 69.5 \\
GPT-5.4 & 83.1\,[82,84] & 86.4 & 76.6 & 94.4 & 75.1 \\
DeepSeek-V4-Pro & 83.1\,[82,84] & 88.2 & 86.1 & 88.5 & 69.5 \\
GLM-5.1 & 81.7\,[80,84] & 83.2 & 89.5 & 79.2 & 75.0 \\
Safe-InnoSpark & 77.0\,[76,78] & 68.0 & 87.6 & 87.3 & 65.2 \\
InnoSpark-235B & 76.4\,[75,78] & 74.2 & 78.0 & 87.5 & 65.9 \\
\bottomrule
\end{tabular}

\caption{The ELBench leaderboard, reporting the overall score with its 95\% bootstrap CI and the four module scores (\%).}
\label{tab:leaderboard}
\end{table*}

\paragraph{Per-module summary.}
The module leaders differ, and so does the distribution of scores within each module. General Capability is led by Gemini-3.5-Flash (93.4) and Claude-Opus-4.8 (91.9), with scores decreasing to the two education-specialized models (74.2 and 68.0). Safety and Trustworthiness is led by the Chinese-developed general models, which hold the top of the module while the three U.S.-developed models are in the bottom four, alongside the education-specialized InnoSpark-235B. It has the widest spread of any module, 19.5 points among the overall-tied leaders, and is the most discriminative module in the suite. Basic Education groups the frontier general models into a high cluster, Gemini-3.5-Flash (94.6), GPT-5.4 (94.4), and Claude-Opus-4.8 (92.7), with the rest within roughly ten points. High-Level Cultivation is the lowest-scoring module overall, with no model exceeding 75.3, and is led by Gemini-3.5-Flash (75.3) and GPT-5.4 (75.1); the education-specialized models rank last.

\begin{figure}[tb]
\centering
\includegraphics[width=0.95\columnwidth]{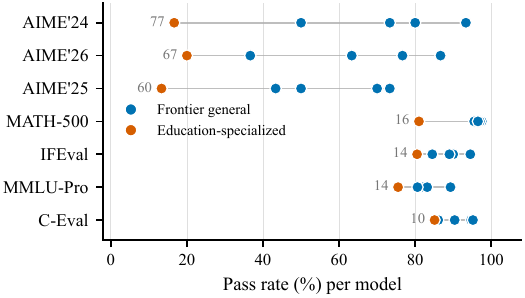}
\caption{General Capability by component, one row per task type (sorted by per-model spread; the number at left is the max$-$min spread). Points are models, colored by type.}
\label{fig:general}
\end{figure}

\paragraph{General Capability by component.}
Within General Capability, the spread between models concentrates in a small number of components (Figure~\ref{fig:general}). On the easier components scores are near ceiling and similar across models, with instruction following close to ceiling for the frontier systems and the curated-knowledge components (MMLU-Pro, C-Eval) in the high eighties and nineties. The competition-mathematics components produce the widest spread. On AIME (2024--2026), per-year success ranges from above $90\%$ for the strongest models to the thirties and fifties for others on the same problems, a wider range than any other general component. The aggregate General Capability score is therefore dominated by competition mathematics, so two models can differ by roughly ten points almost entirely because of AIME. Models fail AIME by losing the reasoning thread across many steps, and sustaining that reasoning is the competence a model needs to explain a difficult problem. The education-specialized InnoSpark-235B shows the same shape, scoring near the top on curated knowledge ($\sim\!85\%$ on C-Eval) but in the teens on AIME, so its deficit is localized to multi-step competition reasoning. A representative per-component table is in Appendix~\ref{app:general}, and the complete per-model breakdown is released with the benchmark.

\begin{figure}[tb]
\centering
\includegraphics[width=0.92\columnwidth]{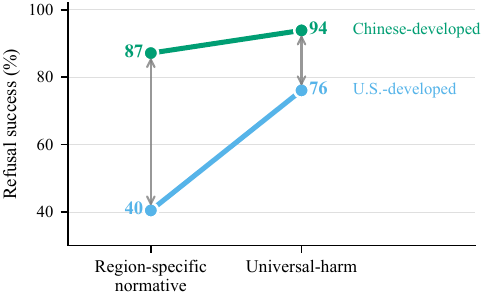}
\caption{Refusal success by model group on the two refusal-category subsets, region-specific normative content and universal-harm content.}
\label{fig:locale}
\end{figure}

\paragraph{Safety by category.}
Within the safety module, the group gap is largest on the refusal task (Figure~\ref{fig:locale}). The U.S.-developed systems decline $39.5$ to $50.4\%$ of requests that should be refused, while three of the four Chinese-developed general models decline $94.7$ to $99.0\%$ and the safety-specialized variant $99.7\%$ (Doubao-Seed-2.0-Pro is an exception at $61.1\%$). Splitting refusal into region-specific normative content and universal-harm content, the group gap is much larger on the region-specific subset than on the universal-harm subset, a difference-in-differences of $28.9$ points ($95\%$ CI $[18.8, 38.5]$). This pattern suggests the gap reflects where the two groups concentrate their safety effort rather than a uniform difference in safety ability. On universal-harm content, where higher refusal is desirable across deployments, the Chinese-developed models still refuse more; on region-specific content, a higher refusal rate measures conformance to a particular jurisdiction's specification, so whether it is desirable depends on the deployment context. The per-category breakdown is in Appendix~\ref{app:safety}.

\begin{figure}[t]
\centering
\includegraphics[width=0.92\columnwidth]{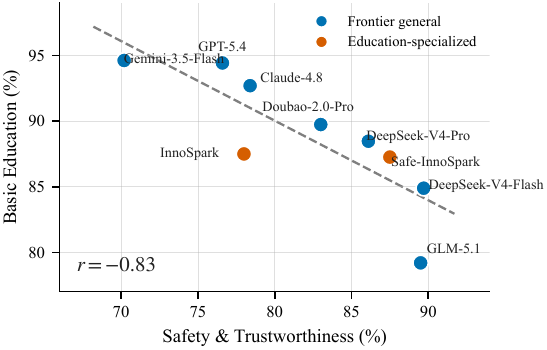}
\caption{Safety against Basic Education across the nine models, which are strongly anti-correlated ($r=-0.83$).}
\label{fig:tradeoff}
\end{figure}

\paragraph{Safety and teaching trade off.}
Across the nine models, Basic Education is strongly anti-correlated with Safety ($r=-0.83$, Spearman $-0.88$; Figure~\ref{fig:tradeoff}), and the correlation is stable under leave-one-model-out recomputation ($[-0.88, -0.79]$), so no single model drives it. Because both modules use tasks the models can perform, this is not a difficulty artifact, and refusal training appears to reduce the openness that practical teaching rewards. The two requirements behave as competing objectives, so a deployment needing both cannot be served by a single education score. High-Level Cultivation is every model's lowest-scoring module and does not separate the field. On the structured judgment task, which is scored by exact reference match without an LLM judge, the models share a systematic error, converging on the same non-reference option on many items, which is why the module is uniformly low and undiscriminating; we analyze this shared blind spot in the supplementary material. This module correlates with General Capability ($r=0.69$), yet the strongest general models do not pull ahead, so scaling general ability alone does not resolve it.

\section{Discussion}

\subsection{The Return on Education-Specific Specialization}
\label{sec:specialization}
\begin{figure}[tb]
\centering
\includegraphics[width=0.92\columnwidth]{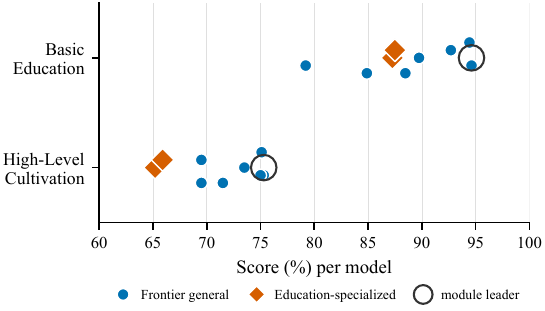}
\caption{The two education modules by model, with the education-specialized models marked.}
\label{fig:vertical}
\end{figure}

The two education-specialized models lead neither education module (Figure~\ref{fig:vertical}), scoring in the middle of the set on Basic Education and at the bottom on High-Level Cultivation, behind general systems that received no education-specific post-training. On this model set the variation attributable to education specialization is small relative to the variation in general capability (Section~\ref{sec:results}). We read this only for what it implies about model development.

The models we evaluate are, to our knowledge, among the strongest education-oriented systems currently available, built by post-training a large general base. Yet within months of their release, general models had reached or exceeded their education scores through ordinary version updates alone. Domain specialization has paid off most durably where the target carries a verifiable reward signal, as in competition mathematics, code, or clinical diagnosis, where a standard answer makes correctness cheap to check and lets post-training improve against a well-defined target \citep{singhal2023clinical,gururangan2020dapt}. The part of education that matters most here has no such signal. High-level educational judgment has no agreed definition of the right response and no reward model to optimize against, so general pre-training and present-day education post-training converge on a similar judgment tendency, the style-over-fit substitution documented in Appendix~\ref{app:cultivation}. This is consistent with all nine models clustering at a similar, modest level on that module and with neither general pre-training nor present-day domain post-training pulling ahead on it. Where much of the domain's instructional content is public and already in the pre-training corpus, a stronger general base may absorb most of what specialization was meant to add. Whether, under these conditions, a separately trained education model retains an advantage over the next general base is the question these results leave open, and ELBench gives that question a measurable form.

\subsection{Toward the Next Generation of Education Benchmarks}
\label{sec:nextgen}
The missing reward signal for educational judgment points to a limit of the current evaluation paradigm, not only of the models. As models advance, static single-turn question answering reaches a construct-validity ceiling for measuring teaching. A recent review of $445$ benchmarks finds that most do not measure the constructs they name \citep{bean2025construct}, and saturation together with pre-training contamination further erodes the discriminative power of fixed test sets \citep{chen2025staticdynamic}. These pressures are sharpest for pedagogy, which is interactive, adaptive, and longitudinal. Strong problem solvers are often weak tutors that reveal answers early \citep{macina2023mathdial}, and teaching quality correlates poorly with model scale or general reasoning \citep{educationq2025}. A rubric applied once to a transcript scores the form of a pedagogical move but not its effect on a learner, the same gap that leaves high-level educational judgment without a reliable signal. Emerging interactive protocols point a way forward, including simulated-student dialogue \citep{educationq2025}, adaptive student personas \citep{jin2025teachtune}, and outcome-grounded scoring of learning gains \citep{scarlatos2025tutor}. Because simulated learners remain imperfect proxies, the next generation of benchmarks will likely pair a contamination-resistant static core like ELBench with a learner-in-the-loop layer.

\section{Conclusion}
ELBench has limitations that frame these results. Module sizes are uneven by design; open-ended scoring depends on rubric judging, which we calibrate but which remains imperfect (Appendices~\ref{app:judge} and~\ref{app:annotation}); the self-built data is expert-verified but synthetic, and the benchmark is text-only; the model set is a sample of nine representative systems, so group-level claims describe this set; and the safety module measures behavior against one education-oriented specification that includes region-specific content, so its result is read within that deployment context (Section~\ref{sec:results}, Appendix~\ref{app:safety}).

We present ELBench, a four-module benchmark that evaluates education-facing LLMs on capability, safety, basic teaching, and high-level cultivation together. Across nine models the module-level profile proves more informative than an aggregate rank, surfacing a near-tie at the top, a safety advantage for the Chinese-developed models that concentrates on region-specific content, and education-specialized models that lead neither education module. ELBench provides a deployment-oriented instrument that makes these development-relevant questions measurable.

\section*{Acknowledgments}
We are grateful to Jiaye Ge of the Shanghai AI Lab for initial project coordination and valuable insights regarding the roadmap. This work was supported by the Shanghai Municipal Education Commission's Special Fund for Educational Large Models (93600-515100-25001).

\bibliography{references}

\begin{thebibliography}{63}
\providecommand{\natexlab}[1]{#1}

\bibitem[{{Anthropic}(2026)}]{anthropic2024claude}
{Anthropic}. 2026.
\newblock {Claude} {Opus} 4.8 System Card.
\newblock System card, Anthropic.
\newblock \url{https://www.anthropic.com/news/claude-opus-4-8}.

\bibitem[{Bai et~al.(2022)Bai, Kadavath, Kundu, Askell
  et~al.}]{bai2022constitutional}
Bai, Y.; Kadavath, S.; Kundu, S.; Askell, A.; et~al. 2022.
\newblock Constitutional AI: Harmlessness from AI Feedback.
\newblock \emph{arXiv:2212.08073}.

\bibitem[{Bean et~al.(2025)Bean, Kearns, Romanou, Hafner, Mayne, Batzner,
  Foroutan, Schmitz, Korgul, Batra, Deb, Beharry, Emde, Foster, Gausen,
  Grandury, Han, Hofmann, Ibrahim, Kim, Kirk, Lin, Liu, Luettgau, Magomere,
  Rystr{\o}m, Sotnikova, Yang, Zhao, Bibi, Bosselut, Clark, Cohan, Foerster,
  Gal, Hale, Raji, Summerfield, Torr, Ududec, Rocher, and
  Mahdi}]{bean2025construct}
Bean, A.~M.; Kearns, R.~O.; Romanou, A.; Hafner, F.~S.; Mayne, H.; Batzner, J.;
  Foroutan, N.; Schmitz, C.; Korgul, K.; Batra, H.; Deb, O.; Beharry, E.; Emde,
  C.; Foster, T.; Gausen, A.; Grandury, M.; Han, S.; Hofmann, V.; Ibrahim, L.;
  Kim, H.; Kirk, H.~R.; Lin, F.; Liu, G. K.-M.; Luettgau, L.; Magomere, J.;
  Rystr{\o}m, J.; Sotnikova, A.; Yang, Y.; Zhao, Y.; Bibi, A.; Bosselut, A.;
  Clark, R.; Cohan, A.; Foerster, J.; Gal, Y.; Hale, S.~A.; Raji, I.~D.;
  Summerfield, C.; Torr, P. H.~S.; Ududec, C.; Rocher, L.; and Mahdi, A. 2025.
\newblock Measuring {W}hat {M}atters: Construct Validity in Large Language
  Model Benchmarks.
\newblock In \emph{NeurIPS Datasets \& Benchmarks}.
\newblock ArXiv:2511.04703.

\bibitem[{Biesta(2015)}]{biesta2015judgement}
Biesta, G. 2015.
\newblock What is Education For? On Good Education, Teacher Judgement, and
  Educational Professionalism.
\newblock \emph{European Journal of Education}, 50(1): 75--87.

\bibitem[{Black and Wiliam(1998)}]{black1998assessment}
Black, P.; and Wiliam, D. 1998.
\newblock Assessment and Classroom Learning.
\newblock \emph{Assessment in Education: Principles, Policy \& Practice}, 5(1):
  7--74.

\bibitem[{Bloom(1984)}]{bloom1984twosigma}
Bloom, B.~S. 1984.
\newblock The 2 Sigma Problem: The Search for Methods of Group Instruction as
  Effective as One-to-One Tutoring.
\newblock \emph{Educational Researcher}, 13(6): 4--16.

\bibitem[{{ByteDance Seed}(2026)}]{bytedance2025seed}
{ByteDance Seed}. 2026.
\newblock {Doubao}-{Seed}-2.0.
\newblock ByteDance Seed Blog.
\newblock \url{https://seed.bytedance.com/en/blog/seed-2-0-official-launch}.

\bibitem[{Chen et~al.(2025)Chen, Chen, Li, Jiang, Wan, He, Ran, Gu, Li, Xie,
  and Ray}]{chen2025staticdynamic}
Chen, S.; Chen, Y.; Li, Z.; Jiang, Y.; Wan, Z.; He, Y.; Ran, D.; Gu, T.; Li,
  H.; Xie, T.; and Ray, B. 2025.
\newblock Benchmarking Large Language Models Under Data Contamination: A Survey
  from Static to Dynamic Evaluation.
\newblock In \emph{EMNLP}, 10080--10098.

\bibitem[{Cohen(1960)}]{cohen1960kappa}
Cohen, J. 1960.
\newblock A Coefficient of Agreement for Nominal Scales.
\newblock \emph{Educational and Psychological Measurement}, 20(1): 37--46.

\bibitem[{{DeepSeek-AI}(2026)}]{deepseek2024v3}
{DeepSeek-AI}. 2026.
\newblock {DeepSeek}-V4: Towards Highly Efficient Million-Token Context
  Intelligence.
\newblock Technical report, DeepSeek-AI.
\newblock \url{https://huggingface.co/deepseek-ai/DeepSeek-V4-Pro}.

\bibitem[{Gehman et~al.(2020)Gehman, Gururangan, Sap, Choi, and
  Smith}]{gehman2020realtoxicity}
Gehman, S.; Gururangan, S.; Sap, M.; Choi, Y.; and Smith, N.~A. 2020.
\newblock RealToxicityPrompts: Evaluating Neural Toxic Degeneration in Language
  Models.
\newblock In \emph{Findings of EMNLP}.

\bibitem[{{Gemini Team, Google}(2026)}]{gemini2023}
{Gemini Team, Google}. 2026.
\newblock {Gemini} 3.5 {Flash}.
\newblock Google.
\newblock
  \url{https://blog.google/innovation-and-ai/models-and-research/gemini-models/gemini-3-5/}.

\bibitem[{{GLM-5 Team, Z.ai}(2026)}]{glm2024chatglm}
{GLM-5 Team, Z.ai}. 2026.
\newblock {GLM}-5.1.
\newblock Z.ai.
\newblock \url{https://docs.z.ai/guides/llm/glm-5.1}.

\bibitem[{Gururangan et~al.(2020)Gururangan, Marasovi\'c, Swayamdipta, Lo,
  Beltagy, Downey, and Smith}]{gururangan2020dapt}
Gururangan, S.; Marasovi\'c, A.; Swayamdipta, S.; Lo, K.; Beltagy, I.; Downey,
  D.; and Smith, N.~A. 2020.
\newblock Don't Stop Pretraining: Adapt Language Models to Domains and Tasks.
\newblock In \emph{ACL}.

\bibitem[{Hendrycks et~al.(2021{\natexlab{a}})Hendrycks, Burns, Basart, Zou,
  Mazeika, Song, and Steinhardt}]{hendrycks2021mmlu}
Hendrycks, D.; Burns, C.; Basart, S.; Zou, A.; Mazeika, M.; Song, D.; and
  Steinhardt, J. 2021{\natexlab{a}}.
\newblock Measuring Massive Multitask Language Understanding.
\newblock In \emph{ICLR}.

\bibitem[{Hendrycks et~al.(2021{\natexlab{b}})Hendrycks, Burns, Kadavath,
  Arora, Basart, Tang, Song, and Steinhardt}]{hendrycks2021math}
Hendrycks, D.; Burns, C.; Kadavath, S.; Arora, A.; Basart, S.; Tang, E.; Song,
  D.; and Steinhardt, J. 2021{\natexlab{b}}.
\newblock Measuring Mathematical Problem Solving With the MATH Dataset.
\newblock In \emph{NeurIPS Datasets \& Benchmarks}.

\bibitem[{Holmes et~al.(2022)Holmes, Porayska-Pomsta, Holstein, Sutherland,
  Baker, Buckingham~Shum, Santos, Rodrigo, Cukurova, Bittencourt, and
  Koedinger}]{holmes2022ethics}
Holmes, W.; Porayska-Pomsta, K.; Holstein, K.; Sutherland, E.; Baker, T.;
  Buckingham~Shum, S.; Santos, O.~C.; Rodrigo, M.~T.; Cukurova, M.;
  Bittencourt, I.~I.; and Koedinger, K.~R. 2022.
\newblock Ethics of {AI} in Education: Towards a Community-Wide Framework.
\newblock \emph{International Journal of Artificial Intelligence in Education},
  32(3): 504--526.

\bibitem[{Huang et~al.(2023)Huang, Bai, Zhu, Zhang, Zhang, Su, Liu, Lv, Zhang,
  Lei, Fu, Sun, and He}]{huang2023ceval}
Huang, Y.; Bai, Y.; Zhu, Z.; Zhang, J.; Zhang, J.; Su, T.; Liu, J.; Lv, C.;
  Zhang, Y.; Lei, J.; Fu, Y.; Sun, M.; and He, J. 2023.
\newblock C-Eval: A Multi-Level Multi-Discipline Chinese Evaluation Suite for
  Foundation Models.
\newblock In \emph{NeurIPS Datasets \& Benchmarks}.

\bibitem[{Jiang et~al.(2026)Jiang, Zhang, Yin, Jin, Lu, Ying, Yu, and
  Kong}]{jiang2025eduguardbench}
Jiang, Y.; Zhang, M.; Yin, X.; Jin, S.; Lu, S.; Ying, Z.; Yu, Z.; and Kong, X.
  2026.
\newblock EduGuardBench: A Holistic Benchmark for Evaluating the Pedagogical
  Fidelity and Adversarial Safety of {LLMs} as Simulated Teachers.
\newblock In \emph{AAAI}.
\newblock ArXiv:2511.06890.

\bibitem[{Jin et~al.(2025)}]{jin2025teachtune}
Jin, H.; et~al. 2025.
\newblock TeachTune: Reviewing Pedagogical Agents Against Diverse Student
  Profiles with Simulated Students.
\newblock In \emph{CHI}.

\bibitem[{Kasneci et~al.(2023)Kasneci, Se{\ss}ler, K\"uchemann, others, and
  Kasneci}]{kasneci2023chatgpt}
Kasneci, E.; Se{\ss}ler, K.; K\"uchemann, S.; others; and Kasneci, G. 2023.
\newblock ChatGPT for Good? On Opportunities and Challenges of Large Language
  Models for Education.
\newblock \emph{Learning and Individual Differences}, 103: 102274.

\bibitem[{{Kimi Team}(2026)}]{kimi2025k15}
{Kimi Team}. 2026.
\newblock {Kimi} {K2.6}.
\newblock Moonshot AI.
\newblock \url{https://www.kimi.com/blog/kimi-k2-6}.

\bibitem[{Lai et~al.(2026)Lai, Xu, Yang et~al.}]{minimax2025}
Lai, X.; Xu, W.; Yang, Y.; et~al. 2026.
\newblock {MiniMax} Sparse Attention.
\newblock \emph{arXiv:2606.13392}.
\newblock MiniMax-M3. \url{https://arxiv.org/abs/2606.13392}.

\bibitem[{Lee et~al.(2026)}]{lee2026openlearnlm}
Lee, U.; et~al. 2026.
\newblock OpenLearnLM Benchmark: A Unified Framework for Evaluating Knowledge,
  Skill, and Attitude in Educational Large Language Models.
\newblock \emph{arXiv:2601.13882}.

\bibitem[{Li et~al.(2024)Li, Zhang, Koto, Yang, Zhao, Gong, Duan, and
  Baldwin}]{li2024cmmlu}
Li, H.; Zhang, Y.; Koto, F.; Yang, Y.; Zhao, H.; Gong, Y.; Duan, N.; and
  Baldwin, T. 2024.
\newblock CMMLU: Measuring Massive Multitask Language Understanding in Chinese.
\newblock In \emph{Findings of ACL}.

\bibitem[{Liang et~al.(2023)Liang, Bommasani, Lee et~al.}]{liang2023holistic}
Liang, P.; Bommasani, R.; Lee, T.; et~al. 2023.
\newblock Holistic Evaluation of Language Models.
\newblock \emph{TMLR}.

\bibitem[{Lin, Hilton, and Evans(2022)}]{lin2022truthfulqa}
Lin, S.; Hilton, J.; and Evans, O. 2022.
\newblock TruthfulQA: Measuring How Models Mimic Human Falsehoods.
\newblock In \emph{ACL}.

\bibitem[{Liu et~al.(2023)Liu, Iter, Xu, Wang, Xu, and Zhu}]{liu2023geval}
Liu, Y.; Iter, D.; Xu, Y.; Wang, S.; Xu, R.; and Zhu, C. 2023.
\newblock G-Eval: {NLG} Evaluation Using {GPT}-4 with Better Human Alignment.
\newblock In \emph{EMNLP}.

\bibitem[{Macina et~al.(2023)Macina, Daheim, Chowdhury, Sinha, Kapur, Gurevych,
  and Sachan}]{macina2023mathdial}
Macina, J.; Daheim, N.; Chowdhury, S.~P.; Sinha, T.; Kapur, M.; Gurevych, I.;
  and Sachan, M. 2023.
\newblock MathDial: A Dialogue Tutoring Dataset with Rich Pedagogical
  Properties Grounded in Math Reasoning Problems.
\newblock In \emph{Findings of EMNLP}.

\bibitem[{Maurya et~al.(2025)Maurya, Srivatsa, Petukhova, and
  Kochmar}]{maurya2025mrbench}
Maurya, K.~K.; Srivatsa, K. V.~A.; Petukhova, K.; and Kochmar, E. 2025.
\newblock Unifying {AI} Tutor Evaluation: An Evaluation Taxonomy for
  Pedagogical Ability Assessment of {LLM}-Powered {AI} Tutors.
\newblock In \emph{NAACL}.

\bibitem[{Mazeika et~al.(2024)Mazeika, Phan, Yin, Zou, Wang, Mu, Sakhaee, Li,
  Basart, Li, Forsyth, and Hendrycks}]{mazeika2024harmbench}
Mazeika, M.; Phan, L.; Yin, X.; Zou, A.; Wang, Z.; Mu, N.; Sakhaee, E.; Li, N.;
  Basart, S.; Li, B.; Forsyth, D.; and Hendrycks, D. 2024.
\newblock HarmBench: A Standardized Evaluation Framework for Automated Red
  Teaming and Robust Refusal.
\newblock In \emph{ICML}.

\bibitem[{{Meta AI}(2025)}]{llama2024herd}
{Meta AI}. 2025.
\newblock The {Llama} 4 Herd: The Beginning of a New Era of Natively Multimodal
  {AI} Innovation.
\newblock Meta AI Blog.
\newblock \url{https://ai.meta.com/blog/llama-4-multimodal-intelligence/}.

\bibitem[{Miao et~al.(2021)Miao, Holmes, Huang, and Zhang}]{miao2021ai}
Miao, F.; Holmes, W.; Huang, R.; and Zhang, H. 2021.
\newblock \emph{AI and Education: Guidance for Policy-makers}.
\newblock Paris, France: UNESCO Publishing.

\bibitem[{Mishra and Koehler(2006)}]{mishra2006tpack}
Mishra, P.; and Koehler, M.~J. 2006.
\newblock Technological Pedagogical Content Knowledge: A Framework for Teacher
  Knowledge.
\newblock \emph{Teachers College Record}, 108(6): 1017--1054.

\bibitem[{{OpenAI}(2026)}]{openai2023gpt4}
{OpenAI}. 2026.
\newblock {GPT}-5.4 Thinking System Card.
\newblock System card, OpenAI.
\newblock \url{https://openai.com/index/gpt-5-4-thinking-system-card/}.

\bibitem[{Panickssery, Bowman, and Feng(2024)}]{panickssery2024selfpreference}
Panickssery, A.; Bowman, S.~R.; and Feng, S. 2024.
\newblock {LLM} Evaluators Recognize and Favor Their Own Generations.
\newblock In \emph{NeurIPS}.

\bibitem[{Perez et~al.(2022)Perez, Huang, Song, Cai, Ring, Aslanides, Glaese,
  McAleese, and Irving}]{perez2022redteaming}
Perez, E.; Huang, S.; Song, F.; Cai, T.; Ring, R.; Aslanides, J.; Glaese, A.;
  McAleese, N.; and Irving, G. 2022.
\newblock Red Teaming Language Models with Language Models.
\newblock In \emph{EMNLP}.

\bibitem[{{Qwen Team}(2026)}]{qwen2024qwen25}
{Qwen Team}. 2026.
\newblock {Qwen3.6}.
\newblock Alibaba Qwen.
\newblock \url{https://huggingface.co/Qwen/Qwen3.6-35B-A3B}.

\bibitem[{R\"ottger et~al.(2024)R\"ottger, Kirk, Vidgen, Attanasio, Bianchi,
  and Hovy}]{rottger2024xstest}
R\"ottger, P.; Kirk, H.~R.; Vidgen, B.; Attanasio, G.; Bianchi, F.; and Hovy,
  D. 2024.
\newblock XSTest: A Test Suite for Identifying Exaggerated Safety Behaviours in
  Large Language Models.
\newblock In \emph{NAACL}.

\bibitem[{Scarlatos et~al.(2025)Scarlatos, Liu, Lee, Baraniuk, and
  Lan}]{scarlatos2025tutor}
Scarlatos, A.; Liu, N.; Lee, J.; Baraniuk, R.; and Lan, A. 2025.
\newblock Training {LLM}-Based Tutors to Improve Student Learning Outcomes in
  Dialogues.
\newblock In \emph{AIED}.

\bibitem[{Shi, Liang, and Xu(2025)}]{educationq2025}
Shi, Y.; Liang, R.; and Xu, Y. 2025.
\newblock EducationQ: Evaluating {LLMs}' Teaching Capabilities Through
  Multi-Agent Dialogue Framework.
\newblock In \emph{ACL}.

\bibitem[{Shulman(1986)}]{shulman1986knowledge}
Shulman, L.~S. 1986.
\newblock Those Who Understand: Knowledge Growth in Teaching.
\newblock \emph{Educational Researcher}, 15(2): 4--14.

\bibitem[{Singhal et~al.(2023)Singhal, Azizi, Tu, others, and
  Natarajan}]{singhal2023clinical}
Singhal, K.; Azizi, S.; Tu, T.; others; and Natarajan, V. 2023.
\newblock Large Language Models Encode Clinical Knowledge.
\newblock \emph{Nature}, 620(7972): 172--180.

\bibitem[{Song et~al.(2025)Song, Liu, Lu, Zhang, Liu, Lv, Wang, Zhou, Tan,
  Jiang, and Hao}]{song2025innospark}
Song, S.; Liu, W.; Lu, Y.; Zhang, R.; Liu, T.; Lv, J.; Wang, X.; Zhou, A.; Tan,
  F.; Jiang, B.; and Hao, H. 2025.
\newblock Cultivating Helpful, Personalized, and Creative {AI} Tutors: A
  Framework for Pedagogical Alignment using Reinforcement Learning.
\newblock \emph{arXiv:2507.20335}.

\bibitem[{Sun et~al.(2023)Sun, Zhang, Deng, Cheng, and
  Huang}]{sun2023chinesesafety}
Sun, H.; Zhang, Z.; Deng, J.; Cheng, J.; and Huang, M. 2023.
\newblock Safety Assessment of Chinese Large Language Models.
\newblock \emph{arXiv:2304.10436}.

\bibitem[{Tan et~al.(2025)Tan, Zhuang, Montgomery, Tang, Cuadron, Wang, Popa,
  and Stoica}]{tan2024judgebench}
Tan, S.; Zhuang, S.; Montgomery, K.; Tang, W.~Y.; Cuadron, A.; Wang, C.; Popa,
  R.~A.; and Stoica, I. 2025.
\newblock JudgeBench: A Benchmark for Evaluating {LLM}-Based Judges.
\newblock In \emph{ICLR}.

\bibitem[{van Es and Sherin(2002)}]{vanes2002noticing}
van Es, E.~A.; and Sherin, M.~G. 2002.
\newblock Learning to Notice: Scaffolding New Teachers' Interpretations of
  Classroom Interactions.
\newblock \emph{Journal of Technology and Teacher Education}, 10(4): 571--596.

\bibitem[{VanLehn(2011)}]{vanlehn2011relative}
VanLehn, K. 2011.
\newblock The Relative Effectiveness of Human Tutoring, Intelligent Tutoring
  Systems, and Other Tutoring Systems.
\newblock \emph{Educational Psychologist}, 46(4): 197--221.

\bibitem[{Vygotsky(1978)}]{vygotsky1978mind}
Vygotsky, L.~S. 1978.
\newblock \emph{Mind in Society: The Development of Higher Psychological
  Processes}.
\newblock Harvard University Press.

\bibitem[{Wang et~al.(2019)Wang, Pruksachatkun, Nangia, Singh, Michael, Hill,
  Levy, and Bowman}]{wang2019superglue}
Wang, A.; Pruksachatkun, Y.; Nangia, N.; Singh, A.; Michael, J.; Hill, F.;
  Levy, O.; and Bowman, S.~R. 2019.
\newblock SuperGLUE: A Stickier Benchmark for General-Purpose Language
  Understanding Systems.
\newblock In \emph{NeurIPS}.

\bibitem[{Wang et~al.(2023)Wang, Chen, Pei, others, Song, and
  Li}]{wang2023decodingtrust}
Wang, B.; Chen, W.; Pei, H.; others; Song, D.; and Li, B. 2023.
\newblock DecodingTrust: A Comprehensive Assessment of Trustworthiness in {GPT}
  Models.
\newblock In \emph{NeurIPS Datasets \& Benchmarks}.

\bibitem[{Wang et~al.(2024{\natexlab{a}})Wang, Xu, Li, Zhang, Liang, Tang, Yu,
  and Wen}]{wang2024largeedu}
Wang, S.; Xu, T.; Li, H.; Zhang, C.; Liang, J.; Tang, J.; Yu, P.~S.; and Wen,
  Q. 2024{\natexlab{a}}.
\newblock Large Language Models for Education: A Survey and Outlook.
\newblock \emph{arXiv:2403.18105}.

\bibitem[{Wang et~al.(2024{\natexlab{b}})Wang, Ma, Zhang
  et~al.}]{wang2024mmlupro}
Wang, Y.; Ma, X.; Zhang, G.; et~al. 2024{\natexlab{b}}.
\newblock MMLU-Pro: A More Robust and Challenging Multi-Task Language
  Understanding Benchmark.
\newblock In \emph{NeurIPS Datasets \& Benchmarks}.

\bibitem[{Wei et~al.(2022)Wei, Wang, Schuurmans, Bosma, Ichter, Xia, Chi, Le,
  and Zhou}]{wei2022cot}
Wei, J.; Wang, X.; Schuurmans, D.; Bosma, M.; Ichter, B.; Xia, F.; Chi, E.; Le,
  Q.; and Zhou, D. 2022.
\newblock Chain-of-Thought Prompting Elicits Reasoning in Large Language
  Models.
\newblock In \emph{NeurIPS}.

\bibitem[{Wei et~al.(2025)Wei, Wang, Bi, Chen, Li, Jiang, Lin, Zhang, Song, Li,
  Zhou, and Hao}]{wei2025elmes}
Wei, S.; Wang, X.; Bi, S.; Chen, J.; Li, R.; Jiang, B.; Lin, X.; Zhang, M.;
  Song, Y.; Li, B.; Zhou, A.; and Hao, H. 2025.
\newblock {ELMES}: An Automated Framework for Evaluating Large Language Models
  in Educational Scenarios.
\newblock \emph{arXiv:2507.22947}.

\bibitem[{Weidinger et~al.(2021)Weidinger, Mellor, Rauh
  et~al.}]{weidinger2021ethical}
Weidinger, L.; Mellor, J.; Rauh, M.; et~al. 2021.
\newblock Ethical and Social Risks of Harm from Language Models.
\newblock \emph{arXiv:2112.04359}.

\bibitem[{{xAI}(2025)}]{xai2025grok}
{xAI}. 2025.
\newblock {Grok} 4 Model Card.
\newblock Model card, xAI.
\newblock \url{https://data.x.ai/2025-08-20-grok-4-model-card.pdf}.

\bibitem[{Xu et~al.(2025)Xu, Bai, Sun et~al.}]{xu2025edubench}
Xu, B.; Bai, Y.; Sun, H.; et~al. 2025.
\newblock EduBench: A Comprehensive Benchmarking Dataset for Evaluating Large
  Language Models in Diverse Educational Scenarios.
\newblock \emph{arXiv:2505.16160}.

\bibitem[{Zhang et~al.(2025)}]{zhang2025omniedubench}
Zhang, M.; et~al. 2025.
\newblock OmniEduBench: A Comprehensive Chinese Benchmark for Evaluating Large
  Language Models in Education.
\newblock \emph{arXiv:2510.26422}.

\bibitem[{Zhang et~al.(2024)Zhang, Lei, Wu, Sun, Huang, Long, Liu, Lei, Tang,
  and Huang}]{zhang2024safetybench}
Zhang, Z.; Lei, L.; Wu, L.; Sun, R.; Huang, Y.; Long, C.; Liu, X.; Lei, X.;
  Tang, J.; and Huang, M. 2024.
\newblock SafetyBench: Evaluating the Safety of Large Language Models.
\newblock In \emph{ACL}.

\bibitem[{Zhao et~al.(2026)Zhao, Yu, Yuan, He, and Wen}]{zhao2026shape}
Zhao, S.; Yu, K.; Yuan, Y.; He, P.; and Wen, H. 2026.
\newblock SHAPE: Unifying Safety, Helpfulness and Pedagogy for Educational
  {LLMs}.
\newblock \emph{arXiv:2604.22134}.

\bibitem[{Zheng et~al.(2023)Zheng, Chiang, Sheng, Zhuang, Wu, Zhuang, Lin, Li,
  Li, Xing, Zhang, Gonzalez, and Stoica}]{zheng2023judging}
Zheng, L.; Chiang, W.-L.; Sheng, Y.; Zhuang, S.; Wu, Z.; Zhuang, Y.; Lin, Z.;
  Li, Z.; Li, D.; Xing, E.~P.; Zhang, H.; Gonzalez, J.~E.; and Stoica, I. 2023.
\newblock Judging {LLM}-as-a-Judge with {MT}-Bench and Chatbot Arena.
\newblock In \emph{NeurIPS Datasets \& Benchmarks}.

\bibitem[{Zhou et~al.(2023)Zhou, Lu, Mishra, Brahma, Basu, Luan, Zhou, and
  Hou}]{zhou2023ifeval}
Zhou, J.; Lu, T.; Mishra, S.; Brahma, S.; Basu, S.; Luan, Y.; Zhou, D.; and
  Hou, L. 2023.
\newblock Instruction-Following Evaluation for Large Language Models.
\newblock \emph{arXiv:2311.07911}.

\end{thebibliography}

\appendix

\section{General Capability: Per-Component Breakdown}
\label{app:general}
This appendix expands the General Capability results (Section~\ref{sec:results}). Table~\ref{tab:general-components} reports per-component pass rates for a representative subset of models spanning the score range; the full per-model breakdown is released with the benchmark. The pattern discussed in the body is visible in the table. On the easier components, scores are near ceiling and similar across models: IFEval, MATH-500, and the curated-knowledge components (MMLU-Pro, C-Eval) fall in the high eighties and nineties for the frontier systems. Competition mathematics (AIME 2024--2026) produces the widest spread: a strong general system can score above $90\%$ on AIME in one year while another scores in the thirties to fifties on the same problems. The education-specialized InnoSpark-235B illustrates this pattern. It scores $85.1\%$ on C-Eval and $81.0\%$ on MATH-500 but falls to $13.3$--$20.0\%$ across the three AIME years, so its capability deficit is localized to multi-step competition mathematics and is not a broad capability gap. Because the aggregate General Capability score is dominated by this component, two models can differ by roughly ten points almost entirely on competition mathematics, a component that a tutoring-product deployer and a contest-aid deployer would weight differently.

\begin{table*}[t]
\centering
\small
\begin{tabular}{lccccccc}
\toprule
Model & MMLU-Pro & C-Eval & IFEval & MATH & AIME'24 & AIME'25 & AIME'26 \\
\midrule
Gemini-3.5-Flash    & 89.3 & 94.7 & 94.5 & 95.5 & 50.0 & 43.3 & 36.7 \\
GPT-5.4             & 82.1 & 86.1 & 90.0 & 97.5 & 80.0 & 50.0 & 63.3 \\
DeepSeek-V4-Flash   & 80.6 & 90.4 & 84.5 & 96.5 & 73.3 & 70.0 & 76.7 \\
Doubao-Seed-2.0-Pro & 83.2 & 95.2 & 89.0 & 97.0 & 93.3 & 73.3 & 86.7 \\
InnoSpark-235B      & 75.5 & 85.1 & 80.5 & 81.0 & 16.7 & 13.3 & 20.0 \\
\bottomrule
\end{tabular}
\caption{Per-component pass rates (\%) on General Capability for a representative subset of models spanning the score range. All nine models are evaluated on all components; this table shows five representative models, and the complete per-model breakdown is released with the benchmark. Scores are near ceiling on the easier components and spread widely on AIME, where the education-specialized model scores lowest.}
\label{tab:general-components}
\end{table*}

\section{Safety: Per-Task and Per-Category Breakdown}
\label{app:safety}
This appendix expands the safety results (Section~\ref{sec:results}). Table~\ref{tab:safety-tasks} decomposes the safety module into its five task families. The advantage of the Chinese-developed models is concentrated in the refusal family, where they reach $94.7$--$99.7\%$ against $39.5$--$50.4\%$ for the U.S.-developed systems. The refusal family is also the one that splits into region-specific and universal-harm categories and produces the difference-in-differences reported in the body. The advantage is not uniform across families. Benign answering is near ceiling for every model, so no model over-refuses legitimate questions. On adversarial robustness, the ordering does not follow the refusal ordering. GPT-5.4 reaches $100\%$ and Claude $90.2\%$ despite refusing least, while several Chinese-developed models score between $40\%$ and $70\%$. The safety advantage is therefore concentrated in the refusal and guidance families and does not extend to every safety task. The teaching-safety family is the hardest for all models, with scores between the high thirties and the high fifties. It is an education-specific multi-select task on which no group scores highly, indicating headroom that is independent of the refusal advantage.

\begin{table*}[t]
\centering
\small
\begin{tabular}{lccccc}
\toprule
Model & Refuse & Guide & Benign & Teach-Saf. & Adv. \\
\midrule
DeepSeek-V4-Flash   & 97.0 & 99.3 & 100  & 59.3 & 67.6 \\
GLM-5.1             & 99.0 & 98.5 & 98.4 & 56.7 & 70.5 \\
Safe-InnoSpark      & 99.7 & 99.3 & 99.7 & 37.7 & 72.8 \\
DeepSeek-V4-Pro     & 94.7 & 97.3 & 100  & 58.7 & 43.0 \\
InnoSpark-235B      & 75.4 & 90.0 & 100  & 40.0 & 56.2 \\
Doubao-Seed-2.0-Pro & 61.1 & 100  & 100  & 51.7 & 100  \\
Claude-Opus-4.8     & 50.4 & 98.0 & 100  & 49.0 & 90.2 \\
GPT-5.4             & 39.5 & 100  & 100  & 44.7 & 100  \\
Gemini-3.5-Flash    & 42.0 & 83.9 & 100  & 55.3 & 54.2 \\
\bottomrule
\end{tabular}
\caption{Safety module decomposed into its five task families (\%). The Chinese-developed models' lead concentrates in refusal and safe guidance; benign answering is saturated for all; adversarial robustness does not follow the same ordering.}
\label{tab:safety-tasks}
\end{table*}

\section{High-Level Cultivation: Uniform-Deviation Analysis}
\label{app:cultivation}
This appendix expands the High-Level Cultivation result (Section~\ref{sec:results}). It examines the structured judgment task, which presents 500 four-option items and is scored by exact match to a single expert reference option with no LLM judge. We analyze all ten evaluated models to characterize why the module scores low and does not separate the field.

\paragraph{Uniform deviation.}
Item-level correctness is bimodal rather than uniform: across the ten models, 140 of the 500 items are answered correctly by all ten and 87 are answered correctly by none, with a sparse middle. On the items the field misses, the errors are concordant. On 105 items at least eight of the ten models select a single common non-reference option, on 84 items at least nine do, and on 53 items all ten do; the mean share of models on the common non-reference option is $0.97$. Because the error is shared across models, the task has near-zero discriminative power on this cluster (median item discrimination index $D=0$), which is the mechanism behind the module's low and undifferentiated scores. The per-model to source-item alignment is a validated bijection over the 500 items, and the extracted option agrees with the deterministic grader on 4{,}999 of 5{,}000 model responses, so the target set is not an alignment or parsing artifact.

\paragraph{Mechanism.}
The shared error is a substitution of pedagogical style for pedagogical fit. Models favor the option that is more gentle, more Socratic, more elaborate, or phrased in a more student-centered register over the option that best serves the specific developmental goal an item names. Two worked items illustrate the pattern. In an accountability item (goal: responsibility and commitment), the reference answer states that standing by while a classmate is bullied is itself wrong, and all ten models instead select a softer perspective-taking prompt. In an emotion-expression item (goal: empathy), the reference answer organizes a class-level reflection, and nine of ten models instead select private reassurance. On some items the model rationale first identifies the reference answer as correct and then declines it as too direct, an explicit override of the goal-optimal choice by a generic style preference.

\paragraph{Bias taxonomy.}
Classifying the 105 uniform-deviation items (Table~\ref{tab:cultivation-bias}) yields six recurring biases across two item formats. Classification items, which ask which competency a scenario illustrates, are dominated by surface keyword matching, in which the model selects the competency whose name shares vocabulary with the scenario rather than the functional construct the expert isolates. Best-response items, which ask which teacher reply best promotes a stated goal, are dominated by preferences for the more elaborate, more Socratic or empathetic, or more gently phrased option, and by avoidance of a blunt but correct reference option in favor of a fluent distractor.

\begin{table}[t]
\centering
\small
\begin{tabular}{lc}
\toprule
Systematic bias & Items \\
\midrule
Surface keyword matching (classification) & 47 \\
Elaboration preference & 18 \\
Blunt-reference avoidance & 14 \\
Socratic or empathy over-generalization & 12 \\
Progressive-register preference & 8 \\
Gentleness over accountability & 6 \\
\bottomrule
\end{tabular}
\caption{Systematic biases on the 105 items where at least eight of ten models select the same non-reference option, by the item's dominant bias. Classification items are dominated by keyword matching; best-response items are dominated by preferences for pedagogical surface features over goal fit.}
\label{tab:cultivation-bias}
\end{table}

\section{Data Generation Details}
\label{app:datagen}
This appendix expands the human-in-the-loop (HITL) pipeline of Section~\ref{sec:datagen}, which produced the self-built families: the three general-safety families (refusal, safe guidance, benign answering) and the two high-level-cultivation families (educational judgment, educational generation).

\paragraph{Seed taxonomy.}
Safety seeds are organized by the refusal taxonomy used in the analysis: \emph{region-specific normative} categories (state-governance, core-values, ethnic and religious content) and \emph{universal-harm} categories (privacy, illegal activity, dangerous instructions). Safe-guidance and benign-answering seeds pair each harmful or benign topic with, respectively, the constructive-redirection and the must-answer pattern. High-level-cultivation seeds enumerate classroom-judgment situations (e.g., emotion regulation, growth mindset, caregiver-anxiety) for the structured task and artifact types (scored feedback, corrected explanation) for the generation task.

\paragraph{Generation and screening.}
Seeds are expanded by several state-of-the-art generator LLMs under family-specific meta-prompts that fix the output schema and require each generated item to preserve its seed's core pedagogical or safety conflict; using multiple generators mitigates single-model bias. The raw corpus is then pre-screened automatically for formatting errors, near-duplicates (by semantic-similarity threshold), and rule violations.

\paragraph{Meta-prompt template.}
Each family uses a meta-prompt of the form: \emph{``Given the seed item below and its category label $\langle$\,c\,$\rangle$, generate $k$ new items that preserve the same $\langle$\,conflict type\,$\rangle$ but vary the subject, grade level, and surface form. Return JSON with fields $\langle$\,schema\,$\rangle$. Do not alter the intended correct behavior.''} The exact schema per family and the full list of generator models are released with the benchmark.

\paragraph{Iterative HITL review.}
Annotators with pedagogical and safety expertise cross-review the screened items. For safety families they assess the plausibility and severity of the embedded request, the correctness of the intended behavior label, and the distinctness of options; for high-level-cultivation families they assess the realism of the situation, the correctness of the reference, and the discriminability of the preferable option. Items are refined or discarded over several rounds, and a final expert pass verifies factual accuracy and category labels. Sensitive items are paraphrased or withheld in any public release.

\section{Curation of Reused Benchmarks}
\label{app:curation}
This appendix expands the curation pipeline of Section~\ref{sec:datagen} for the General Capability module, which is assembled from public sources. We hold this pipeline to the same standard as the synthesis pipeline: every item is selected and verified by experts, so that the reused portion is as controlled as the self-built portion.

\paragraph{Candidate pool.}
For each source we start from its public test split: MMLU-Pro and C-Eval for subject knowledge, IFEval for instruction following, and a mathematics ladder of the MATH-500 subset and AIME 2024--2026. Train and validation splits are excluded so that no item is drawn from material commonly used for model training.

\paragraph{Balanced sampling.}
We sample for balanced coverage of each source's internal structure: subject categories for MMLU-Pro and C-Eval, instruction types for IFEval, and difficulty levels for the mathematics sources. This prevents a single subject or difficulty band from dominating a component and keeps the curated set representative of the ability the source measures. The AIME years are kept in full (30 problems each) because the competition set is already small and difficulty-balanced by design.

\paragraph{Quality filtering and de-duplication.}
Experts review the sampled items and discard those that are low quality, ambiguous, or malformed: unclear stems, disputed or non-unique answers, broken options, and formatting damage introduced upstream. Duplicate and near-duplicate items are removed so that no question is counted twice across or within sources.

\paragraph{Answer verification and contamination check.}
For every retained item an expert verifies the reference answer against which the model will be scored, since an incorrect key would silently penalize correct responses. We also check for train-set contamination and, where a contamination-resistant variant exists, prefer it; MMLU-Pro is chosen over MMLU for this reason. Items that fail verification are corrected or discarded.

\begin{table}[t]
\centering
\small
\begin{tabular}{llr}
\toprule
Source & Measures & Items \\
\midrule
MMLU-Pro & Subject knowledge & 196 \\
C-Eval & Chinese-curriculum knowledge & 208 \\
IFEval & Instruction following & 200 \\
MATH-500 & Mathematics (subset) & 200 \\
AIME 2024 & Competition mathematics & 30 \\
AIME 2025 & Competition mathematics & 30 \\
AIME 2026 & Competition mathematics & 30 \\
\midrule
\textbf{Total} & & \textbf{894} \\
\bottomrule
\end{tabular}
\caption{Final item count per source in the curated General Capability module, after balanced sampling, quality filtering, and verification.}
\label{tab:curation-counts}
\end{table}

\section{LLM Judge Selection and Calibration}
\label{app:judge}
Open-ended responses, namely the instructional-quality, safe-redirection, and educational-generation tasks, are scored by an LLM judge selected for high agreement with human annotation, following the practice established for pedagogical evaluation \citep{jiang2025eduguardbench}.

\paragraph{Gold-standard set.}
We sample 300 responses for the gold set, 100 from each of the three open-ended task families (safe redirection, instructional quality on Basic Education, and educational generation on High-Level Cultivation). The sample is stratified to be balanced across evaluated models and across the score range. Each sampled response is independently annotated by three domain experts in a double-blind fashion. We measure inter-annotator agreement with quadratic weighted Cohen's $\kappa$ \citep{cohen1960kappa} averaged over annotator pairs, and obtain $0.884$, $0.862$, and $0.821$ on the three families respectively ($0.856$ mean); this level of human-human agreement establishes a ceiling against which the judge-human agreement below should be read. Final gold labels use the median expert score, with disagreements resolved in consensus review.

\paragraph{Candidate judges and selection.}
We benchmark five candidate judges spanning distinct model families: Qwen3.6 \citep{qwen2024qwen25}, Kimi-2.6 \citep{kimi2025k15}, Grok-4.3 \citep{xai2025grok}, MiniMax-M3 \citep{minimax2025}, and Llama-4 \citep{llama2024herd}. Each candidate scores every gold-set item under the same rubric prompt used in the main evaluation, zero-shot with greedy decoding. For each candidate we report agreement with the human gold labels using quadratic weighted Cohen's $\kappa$, which weights disagreements by the square of their distance on the rating scale and is appropriate for these graded rubric scores. Qwen3.6 attained the highest agreement on every task family and is used as the ELBench judge (Table~\ref{tab:judge}). Its agreement with the human gold labels ($0.83$ mean) approaches the human-human ceiling above, indicating that the judge tracks expert scoring about as closely as experts track one another.

\begin{table}[h]
\centering
\small
\setlength{\tabcolsep}{3pt}
\begin{tabular}{lcccc}
\toprule
Candidate judge & Safe-Redir. & \shortstack{Basic\\Education} & Cultivation & Mean \\
\midrule
Kimi-2.6   & 0.816 & 0.812 & 0.765 & 0.798 \\
Grok-4.3   & 0.791 & 0.784 & 0.742 & 0.772 \\
MiniMax-M3 & 0.804 & 0.798 & 0.751 & 0.784 \\
Llama-4    & 0.773 & 0.761 & 0.718 & 0.751 \\
\textbf{Qwen3.6 (selected)} & \textbf{0.847} & \textbf{0.836} & \textbf{0.792} & \textbf{0.825} \\
\midrule
\textit{Human--human} & 0.884 & 0.862 & 0.821 & 0.856 \\
\bottomrule
\end{tabular}
\caption{Judge-human agreement (quadratic weighted Cohen's $\kappa$) of candidate judges with the human gold labels, per open-ended task family: safe redirection (Safe-Redir.), instructional quality on Basic Education, and educational generation on High-Level Cultivation. Qwen3.6 attains the highest agreement on every family and is used for all open-ended scoring. The bottom row reports inter-annotator (human--human) agreement as an upper reference.}
\label{tab:judge}
\end{table}

\paragraph{Stabilization and bias controls.}
To reduce single-call variance, each open-ended item's final label is the majority over $N=9$ independent judge calls, following the best-of-$N$ voting used in comparable pedagogical evaluation \citep{jiang2025eduguardbench}. Where a judgment depends on presentation order, the order is randomized to control position bias \citep{zheng2023judging}. None of the nine evaluated models belongs to the Qwen family, so same-family self-preference between the judge and an evaluated model \citep{panickssery2024selfpreference} does not apply to our results.

\section{Annotation Governance}
\label{app:annotation}
The expert annotators hold advanced degrees in computer science or education technology and have experience in NLP, AI safety, or educational assessment. Annotation proceeds in three stages: a training and calibration round on shared examples to align on the rubric; independent double-blind annotation of the gold set; and consensus meetings to resolve disagreements. The guidelines specify, per task type, the acceptability criteria and worked examples of acceptable and unacceptable responses; sensitive examples are paraphrased or withheld in any public release. This governance follows the protocol used in comparable education-safety annotation \citep{jiang2025eduguardbench} and supports both the data curation of Appendix~\ref{app:datagen} and the judge selection of Appendix~\ref{app:judge}.

\section{Uncertainty Estimation}
\label{app:bootstrap}
\begin{figure}[t]
\centering
\includegraphics[width=0.92\columnwidth]{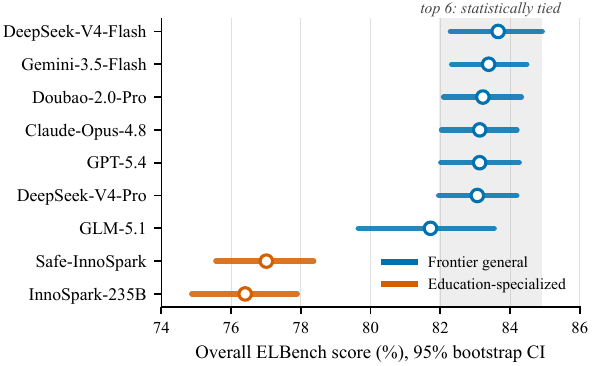}
\caption{Overall score with 95\% bootstrap confidence intervals, colored by model group. The leading intervals overlap, and the two education-specialized models are separated from the band.}
\label{fig:overall}
\end{figure}

The confidence intervals in Table~\ref{tab:leaderboard} and Figure~\ref{fig:overall} are computed by item-level bootstrap with $10{,}000$ resamples. Within each resample we resample items with replacement inside each module, recompute every model's module scores and overall score on the shared resample, and take the $2.5$ and $97.5$ percentiles of each model's bootstrap distribution as its interval. Paired comparisons report the bootstrap probability that one model's overall score exceeds another's across resamples; we treat a pair as distinguishable when this probability exceeds $0.95$. Under this procedure the top six overall scores are mutually indistinguishable, and the two education-specialized models are separated from the leading band ($P\!\approx\!1.0$, disjoint intervals). The safety difference-in-differences in Section~\ref{sec:results} is computed by the same item-level bootstrap applied to the four-cell (group $\times$ category) refusal-success contrast.

\end{document}